\documentclass[11pt,column]{article}
\usepackage{amsfonts,amssymb,amsmath,mathtools,bm,amsthm}
\usepackage{amssymb,amsmath,amsthm,mathrsfs,ulem,tikz, caption}

\title{How Many Factors 
Influence Minima in SGD?}
\author{Victor Luo  and Yazhen Wang \\ Department of Statistics, University of Wisconsin-Madison \\ Madison, WI 53706, USA. \\ Email: vluo@wisc.edu, yzwang@stat.wisc.edu}

\usepackage[margin=0.7in]{geometry}

\newtheorem{result}{Result}[section]

\newtheorem{assump}{Assumption}

\newcommand{\ff}{\mathfrak{f}}

\begin{document}
\maketitle
\begin{abstract}
Stochastic gradient descent (SGD) is often applied to train Deep Neural Networks (DNNs), and 
research efforts have been devoted to  
investigate the convergent dynamics of SGD and minima found by SGD. The influencing factors identified in the literature include learning rate, batch size, Hessian, and gradient covariance, and stochastic differential equations 
are used to model SGD and establish the relationships among these factors for characterizing minima found by SGD. It 
has been found that the ratio of batch size to learning rate is a main factor in highlighting the underlying SGD dynamics; however, the influence of other important factors such as the Hessian and gradient covariance is not 
entirely agreed upon. This paper describes the factors and relationships in the recent literature and presents 
numerical findings on the relationships. In particular, it confirms the four-factor and general relationship results 
obtained in Wang (2019), while the three-factor and associated relationship results found in Jastrz\c{e}bski et al. (2018) may not hold beyond the considered special case.  

\end{abstract}


\section{Introduction} 

With the rise of big data analytics, multi-layer deep neural networks (DNNs) have surfaced as one of the most powerful machine learning methods. Training a deep network requires optimizing an empirical risk or average loss function, typically done using stochastic gradient descent (SGD). There is a large volume of literature on DNNs. This paper focuses on two recent papers, Wang (2019) and Jastrz\c{e}bski et al. (2018), that 
show that the SGD dynamics are captured by the ratio of the learning rate (LR) to batch size (BS) along with other 
factors. We will refer to this ratio as LR/BS. Jastrz\c{e}bski et al. (2018) considered a special setting to identify three factors and established a relationship 
among the three factors to character minima in SGD. Wang (2019) studied a general setting to find four 
factors and derived a general relationship to describe minima in SGD. Moreover, Jastrz\c{e}bski et al. (2018) can be 
treated as a special case of Wang (2019). 
We detail the two approaches and their settings in the following sections, and refer to the specific equations of each paper involving relationships with the LR/BS in the titles. 
Their differences are investigated numerically.

The rest of the paper is organized as follows. Sections 2 and 3 provide brief reviews of Jastrz\c{e}bski et al. (2018) and Wang (2019), respectively. 
Section 4 describes neutral network model setup, 
data sets, and software packages. Section 5 presents numerical results. 

\section{The SGD Behavior under a Special Case in Jastrz\c{e}bski et al. (2018)} 
\label{section2} 
This section provides an overview of the main results in 
Jastrz\c{e}bski et al. (2018) describing the SGD behavior. 
Consider a model parameterized by ${\bm \theta}$ where $\theta_i$ for $i\in \{1,\dots,q\}$ are the components, with $q$ denoting the number of parameters. For $n$ training examples ${\bf x}_j$, $j\in \{1,\dots,n\}$, the average loss function is defined as $L({\bm \theta})=\frac{1}{n} \sum_{j=1}^n l({\bm \theta},{\bf x}_j)$. The corresponding gradient is ${\bf g}({\bm \theta})=\frac{\partial {\bf L}}{\partial {\bm \theta}}$. The gradient is based on the sum over all loss values for all training examples.

Stochastic gradient descent with learning rate $\delta$ is considered, where the update rule is given by

\begin{align} \label{SGD1}
{\bm \theta}_{k+1}={\bm \theta}_k - \delta {\bf g}^{(m)}({\bm \theta}_k),
\end{align}

\noindent where $k$ indexes the discrete update steps, and ${\bf g}^{(m)}$ are the stochastic gradients that arise when considering a minibatch $\mathbb{B}$ of size $m<n$ of random indices drawn uniformly from $\{1,\dots,n\}$. The stochastic gradients ${\bf g}^{(m)}$ form an unbiased estimate of the gradient based in the corresponding subset of training examples ${\bf g}^{(m)}({\bm \theta})=\frac{1}{m} \sum_{j\in \mathbb{B}} \frac{\partial}{\partial {\bm \theta}} l({\bm \theta},{\bf x}_j)$.

If we consider the loss gradient at a randomly chosen data point ${\bf g}_j({\bm \theta})=\frac{\partial}{\partial {\bm \theta}}l({\bm \theta},{\bf x}_j)$ as a random variable induced by the random sampling of the data items, ${\bf g}_j({\bm \theta})$ is an unbiased estimator of the gradient $E[{\bf g}({\bm \theta})]$. This estimator ${\bf g}_j({\bm \theta})$ has finite covariance ${\bf C}({\bm \theta})$ for typical loss functions. With a large enough data set, each item in a batch is a conditional IID sample. 
When the batch size is sufficiently large, ${\bf g}^{(m)}({\bm \theta})$ is the mean of components of the 
form ${\bf g}_j({\bm \theta})$. 
Thus, ${\bf g}^{(m)}({\bm \theta})$ approximately follows a normal distribution with mean close to 
${\bf g}({\bm \theta})$ and covariance ${\bm \Sigma}({\bm \theta})=\frac{1}{m}{\bf C}({\bm \theta})$. Therefore, (\ref{SGD1}) can be rewritten as

\begin{align} \label{SGD2}
{\bm \theta}_{k+1}={\bm \theta}_k -\delta {\bf g}({\bm \theta}_k)+\delta({\bf g}^{(m)}({\bm \theta}_k)-{\bf g}({\bm \theta}_k)),
\end{align}

\noindent where $({\bf g}^{(m)}({\bm \theta}_k)-{\bf g}({\bm \theta}_k))$ is an additive zero mean Gaussian random noise with variance ${\bm \Sigma}(\bf{\theta})=(1/m){\bf C}({\bm \theta})$. Then, (\ref{SGD2}) can be rewritten as

\begin{align} \label{SGD3} 
{\bm \theta}_{k+1}={\bm \theta}_k-\delta {\bf g}({\bm \theta}_k)+\frac{\delta}{\sqrt{m}}{\bm \epsilon} , 
\end{align}

\noindent with ${\bm \epsilon}$ being a zero mean Gaussian random variable with covariance ${\bf C}({\bm \theta})$.

Next, we may model (\ref{SGD3}) by a stochastic differential equation (SDE) of the form 

\begin{align} \label{SDE1}
d{\bm \theta}=-{\bf g}({\bm \theta})dt+\sqrt{\frac{\delta}{m}}{\bf R}({\bm \theta})d{\bf W}(t),
\end{align}

\noindent where ${\bf R}({\bm \theta}){\bf R}({\bm \theta})^T={\bf C}({\bm \theta})$. It is noted that ${\bf R}(\theta)={\bf U}(\theta){\bf D}(\theta)^{1/2}$, and the eigendecomposition of ${\bf C}(\theta)$ is given as ${\bf U}(\theta){\bm \Lambda}(\theta){\bf U}(\theta)^T$, where ${\bm \Lambda}(\theta)$ is the diagonal matrix of eigenvalues and ${\bf U}(\theta)$ is the orthonormal matrix of eigenvectors of ${\bf C}(\theta)$. 

Two assumptions are made:

\begin{assump}
As we expect the training to have arrived in a local minima, the loss surface can be approximated by a quadratic bowl, with minimum at zero loss (reflecting the ability of networks to fully fit the training data). Given this the training can be approximated by an Ornstein-Unhlenbeck process.
\end{assump}

\begin{assump}
Assume ${\bf C}={\bf H}$, that is, the covariance of the gradients and the Hessian of the loss are equal. 
\end{assump}

Under these assumptions, the Hessian is positive definite, and matches the covariance ${\bf C}$. Thus, its eigendecomposition is ${\bf H}={\bf C}={\bf V}{\bm \Lambda} {\bf V}^T$, where ${\bm \Lambda}$ is the diagonal matrix of positive eigenvalues, and ${\bf V}$ is an orthonormal matrix. Let ${\bf z}$ be defined as ${\bf z}={\bf V}^T({\bm \theta}-{\bm \theta}_*)$, where ${\bm \theta}_*$ are the parameters at the minimum.

Now, starting from (\ref{SDE1}), and approximating the average loss $L(\theta)$ as $({\bm \theta}-{\bm \theta}_*)^T {\bf H}({\bm \theta}-{\bm \theta}_*)$, we obtain an Ornstein-Unhlenbeck (OU) process for ${\bf z}$ given as

\begin{align}
d{\bf z}=-{\bm \Lambda}{\bf z}dt+\sqrt{\frac{\delta}{m}}{\bm \Lambda}^{1/2} d{\bf W}(t).
\end{align}

\noindent The stationary distribution of an OU process of this form is Gaussian with mean zero and covariance cov$({\bf z})=\mathbb{E}({\bf z}{\bf z}^T)=\frac{\delta}{2m}{\bf I}$. Furthermore, Jastrz\c{e}bski et al. (2018, Equation 9) 
 obtains the expected loss that can be written as

\begin{align} \label{result1}
\mathbb{E}(L)=\frac{1}{2}\sum_{i=1}^q \lambda_i \mathbb{E}(z_i^2)=\frac{\delta}{4m}\text{Tr}({\bm \Lambda})=\frac{\delta}{4m}\text{Tr}({\bf H}),
\end{align}

\noindent with the expectation being over the stationary distribution of the OU process, and the second equality following from the expression for the OU covariance. We will refer to (\ref{result1}) as Equation\cite{J2018}(9)  for the rest of this paper. We can see from \cite{J2018}(9) that the learning rate to batch size ratio determines the trade-off between width and expected loss associated with SGD dynamics within a minimum centered at a point of zero loss, with $\frac{\mathbb{E}(L)}{\text{Tr}({\bf H})} \propto \frac{\delta}{m}$.

\section{The SGD Behavior under the General Case  in Wang (2019)} 
This section reviews the relevant SGD results in Wang (2019). 
We consider the minimization problem $\min_{\theta \in \Theta} \ff(\theta)$, where the objective function $\ff(\theta)$ is defined on a parameter space $\Theta \subset \mathbb{R}^p$ and assumed to have L-Lipschitz continuous gradients. The plain gradient descent algorithm is defined as $x_k=x_{k-1}-\delta\nabla \ff(x_{k-1})$, where $\nabla$ denotes the gradient operator and $\delta$ is the learning rate. We can model $\{x_k, k=0,1,\dots\}$ by a smooth curve $X(t)$ with ansatz $x_k\approx X(k\delta)$. Define a step function $x_\delta(t)=x_k$ for $k\delta \leq t <(k+1)\delta$, and as $\delta\to 0$, $x_\delta(t)$ approaches $X(t)$ satisfying 

\begin{align} \label{2.3}
\dot{X}(t)+\nabla \ff(X(t))=0,
\end{align}

\noindent where $\dot{X}(t)$ denotes the derivative of $X(t)$ and initial value $X(0)=x_0$. We see that $X(t)$ is a gradient flow associated with the objective function $\ff(\cdot)$.

Now, let $\theta=(\theta_1,\dots,\theta_p)'$ be parameters that we are interested in, and $U$ a relevant random element on a probability space with a given distribution $Q$. Consider an objective function $\ell(\theta;u)$ and its corresponding expectation $E[\ell(\theta;U)]=\ff(\theta)$. We can treat $\ell(\theta;u)$ as a loss function and $\ff(\theta)=E[\ell(\theta;U)]$ its corresponding risk.  Because $g(\theta)$ is usually unavailable, we consider the stochastic optimization problem $\min_{\theta \in \Theta} \mathbb{L}^n (\theta;{\bf U}_n)$, where $\mathbb{L}^n(\theta;{\bf U}_n)=\frac{1}{n} \sum_{i=1}^n \ell(\theta;U_i)$, ${\bf U}_n=(U_1,\dots,U_n)'$ is a sample, and we assume that $U_1,\dots,U_n$ are iid and follow the distribution $Q$. 

We make some smoothing assumptions on the loss function $\ell(\theta; u)$, and 
$E[\ell(\theta;U)]=\ff(\theta)$, $E[\nabla \ell(\theta;U)]=\nabla \ff(\theta)$, $E[\mathbb{H} \ell(\theta;U)]=\mathbb{H} \ff(\theta)$, 
where $\nabla$ is the gradient operator (the first order partial derivatives), and $\mathbb{H}$ is the Hessian operator (the second order partial derivatives).
We further assume that 
$\sqrt{n}[\nabla \mathbb{L}^n (\theta;{\bf U}_n)-\nabla \ff(\theta)]=\frac{1}{\sqrt{n}}\sum_{i=1}^n [\nabla \ell(\theta;U_i)-\nabla \ff(\theta)]$ weakly converges to 
a Gaussian process with mean zero and autocovariance ${\bm \varsigma}(\theta,\upsilon)$ defined below.

Define the cross auto-covariance ${\bm \varsigma}(\theta,\upsilon)=(\varsigma_{ij}(\theta,\upsilon))_{1\leq i,j\leq p}$, $\theta,\upsilon \in \Theta$, where Cov$[\frac{\partial}{\partial \theta_i} \ell(\theta;U),\frac{\partial}{\partial \upsilon_j} \ell(\upsilon;U)]$$=\varsigma_{ij}(\theta,\upsilon)$ are assumed to be continuously differentiable, and L-Lipschitz. Let $\sigma_{ij}(\theta)=$Cov$[\frac{\partial}{\partial \theta_i} \ell(\theta;U),\frac{\partial}{\partial \theta_j} \ell(\theta;U)]=\varsigma_{ij}(\theta,\theta)$, and ${\bm \sigma}^2(\theta)=\text{Var}[\nabla \ell(\theta;U)]=(\sigma_{ij}(\theta))_{1\leq i,j\leq p}={\bm \varsigma}(\theta,\theta)$ is positive definite.


The stochastic gradient scheme can then be viewed as

\begin{align}\label{SGD3}
x_k^m=x_{k-1}^m - \delta \nabla \hat{\mathbb{L}}^m(x_{k-1}^m ; {\bf U}_{mk}^*),
\end{align}

\noindent where ${\bf U}_{mk}^*=(U_{1k}^*,\dots,U_{mk}^*)'$, $k=1,2,\dots$, are independent mini-batches. 


As an analog of gradient flow given by ODE (\ref{2.3}) for the gradient descent algorithm, a continuous-time process $X_\delta^m(t)$ is derived 
to approximate the stochastic gradient descent algorithm by the following stochastic differential equation, 

\begin{align} \label{4.21}
dX_\delta^m(t)=-\nabla \ff(X_\delta^m(t))dt-(\delta/m)^{1/2}{\bm \sigma}(X_\delta^m(t))d{\bf B}(t).
\end{align}



Let $V_\delta^m(t)=(m/\delta)^{1/2}[X_\delta^m(t)-X(t)]$,  
and treat $V_\delta^m$ as random elements in $C([0,T])$, where $C([0,T])$ is the space of all continuous functions on $[0,T]$ with the uniform metric $\max\{|b_1(t)-b_2(t)|:t\in [0,T]\}$ between functions $b_1(t)$ and $b_2(t)$. 
Similarly, we use the SGD iterate $x_k^m$ from (\ref{SGD3}) to define an empirical process $x_\delta^m(t)$. 
Set $v_\delta^m(t)=(m/\delta)^{1/2}[x_\delta^m(t)-X(t)]$, and treat $v_\delta^m$ as random elements in $D([0,T])$, 
where $D([0, T])$ is the Skorokhod space of all c\'adl\'ag functions on $[0, T]$, equipped with the Skorokhod 
metric. 
Wang (2019) established the weak convergence limit of $V_\delta^m(t)$ and $v_\delta^m(t)$ as follows. 

\begin{result}\label{4.2} (Gradient flow central limit theorem in Wang (2019).) 
As $\delta\to 0$ and $m\to \infty$, 
$V_\delta^m(t)$ and $v^m_\delta(t)$, $t\in [0,T]$, weakly converge to $V(t)$ which is a time-dependent Ornstein-Uhlenbeck process satisfying
\begin{align}\label{4.24}
dV(t)=-[\mathbb{H}\ff(X(t))]V(t)dt-{\bm \sigma}(X(t))d{\bf B}(t), V(0)=0, 
\end{align}
where  $\mathbb{H}$ denotes the Hessian operator. 
\end{result}



Suppose that stochastic gradient processes converge to the critical point $\check{\theta}$. 
From a limiting distribution point of view, Result \ref{4.2} indicates that the SGD iterate $x^m_k$ from (\ref{SGD3}) 
and the continuous processes $X_{\delta}^m(t)$ 
generated from the SDE (\ref{4.21}) are 
asymptotically the same as the deterministic solution $X(t)$ of the ordinary differential equation (\ref{2.3}) plus $(\delta/m)^{1/2} V(t)$, where $V(t)$ is the solution of the SDE (\ref{4.24}). The limiting process $V(t)$ is a time-dependent Ornstein-Uhlenbeck process given by (\ref{4.24}). Then, we have the following theorem for the behaviors of 
$\ff(X_\delta^m(t))$ and $\ff(x_\delta^m(t))$
around the critical point $\check{\theta}$.

\begin{result} (Minima in SGD in Wang (2019).) 
Suppose 
the gradient descent process $X(t)$ given by the ordinary differential equation (\ref{2.3}) converges to a critical point $\check{\theta}$ of $\ff(\cdot)$. If $\check{\theta}=X(\infty)$ is a local minimizer with positive definite $\mathbb{H} \ff(\check{\theta})$, then as $t\to \infty$, $V(t)$ has a limiting stationary distribution with mean zero and covariance matrix $\Gamma(\infty)$ satisfying the following algebraic Ricatti equation, 
\begin{equation}
\Gamma(\infty) \mathbb{H}\ff(X(\infty))  + \mathbb{H}\ff(X(\infty))  \Gamma(\infty)     = {\bm \sigma}^2(X(\infty) ) ;
\end{equation} 
moreover, we have 
\begin{align} \label{SGD-minima-1}
E[\ff(X_\delta^m(t))] &=\ff(X(t))+\frac{\delta}{4m}tr[{\bm \sigma}^2(X(\infty))]+o(\delta/m), \\
E[|\nabla \ff(X_\delta^m(t))|^2] &=|\nabla \ff(X(t))|^2+\frac{\delta}{2m}tr[{\bm \sigma}^2(X(\infty)) \mathbb{H}\ff(X(\infty))]+o(\delta/m).  \label{SGD-minima-2}
\end{align}

\noindent If $\check{\theta}$ is a saddle point, $V(t)$ diverges and thus does not have any limiting distribution.
\end{result}

\noindent 
Since the above equations (\ref{SGD-minima-1}) and  (\ref{SGD-minima-2}) correspond to Equations (4.45) and (4.49) 
in Wang (2019), respectively, we will refer to them as Equations \cite{W2019}(4.45) and \cite{W2019}(4.46) for the rest of the paper. Note that under Assumption 2  in Section \ref{section2}, $ {\bm \sigma}^2(X(\infty)) = \mathbb{H}\ff(X(\infty))$; thus, 
(\ref{result1}) can be recovered from (\ref{SGD-minima-1}). However, without the assumption, there may have a  
significant difference between (\ref{result1}) and (\ref{SGD-minima-1})---namely, Equation \cite{W2019}(4.45) can significantly differ from Equation \cite{J2018}(9). 

\section{The Setup for the Numerical Study} 

The numerical analysis in the rest of the paper was conducted in $R$ coupled with Python using packages keras and tensorflow. Keras is a deep learning API written in Python, with runs on top of tensorflow. Tensorflow is a symbolic math library that is used for machine learning applications, especially neural networks. We specifically utilize the GPU version of the packages, with a NVIDIA GeForce GTX 1050 Ti being the graphics card.

\begin{figure}[h]
\caption{ResNet code utilized in $R$.}
\label{ResNet}
\centering
\includegraphics[scale=0.55]{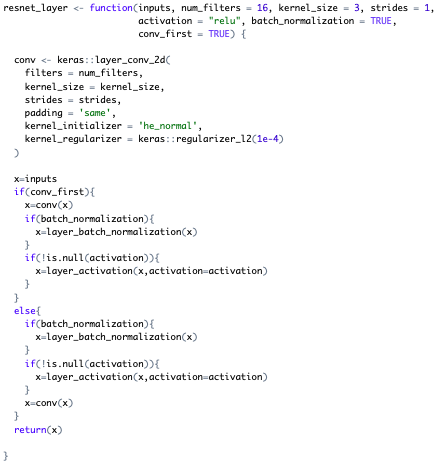}
\includegraphics[scale=0.55]{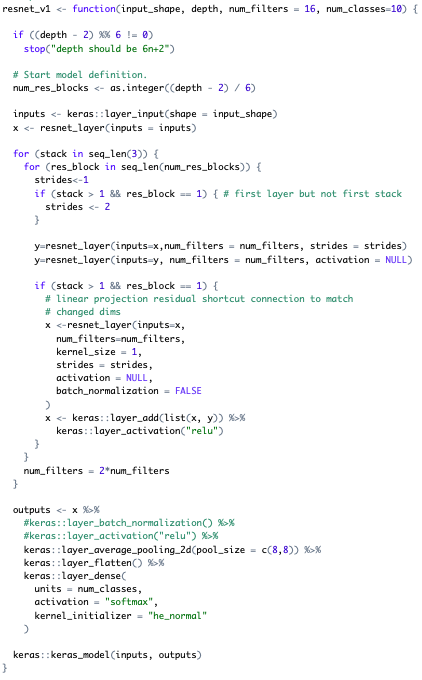}
\end{figure}

The model considered was ResNet 56, where 56 represents the depth of the model. ResNet stacks convolution layers, but introduces the key idea of identity shortcut connections, allowing the model to skip one or more layers. Using a residual block allows the stacked layers to fit an easier residual mapping rather than directly fitting the desired underlaying mapping. The code is given in Figure \ref{ResNet}.

The data set considered was CIFAR-10. CIFAR-10 is a dataset consisting of 60,000 32x32 color images in 10 classes (airplane, automobile, bird, cat, deer, dog, frog, horse, ship, and truck), with 6,000 images per class. The data set is divided in to 50,000 training images and 10,000 test images. The model is fit on the training images and validated using the test images. The training is run on 125 epochs, and the batch size and learning rate will be varied. In addition, the learning rate is reduced by a factor of 10 at checkpoints 60 epochs and 100 epochs to help prevent the loss from plateauing. SGD and batch normalization are used as well. We will explore the Kullback-Leibler divergence loss, which is given as $\sum y_{true}*\log(\frac{y_{true}}{y_{pred}})$ in tensorflow, and the gradient of the standard logistic loss (difference in $y_{true}$ and $y_{pred}$), where $y_{true}$ is the true value and $y_{pred}$ is the predicted from the model. The log loss function was explored in Wang (2019). 

\section{Numerical Results}

In this section, we will first justify Equations \cite{W2019}(4.45) and \cite{W2019}(4.46) through numerical experiment, then highlight the differences between \cite{W2019}(4.45) and \cite{J2018}(9)---namely, the differences between $tr[{\bm \sigma}^2(X(\infty))]$ and $\text{Tr}({\bf H})$, respectively.

\begin{figure}[h!]
\caption{Evolution of Training and Validation Accuracy for varying BS/LR using ResNet-56 on CIFAR-10 with KLD Loss.}
\label{KLDLossGraphs}
\centering
\includegraphics[scale=0.65]{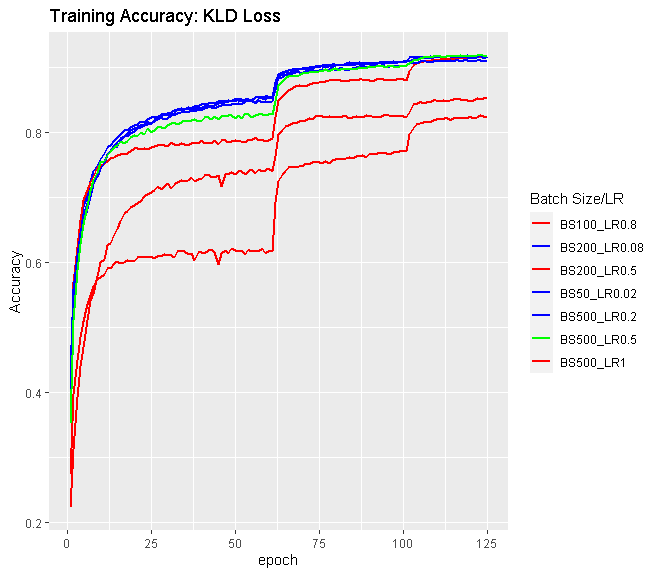}
\includegraphics[scale=0.65]{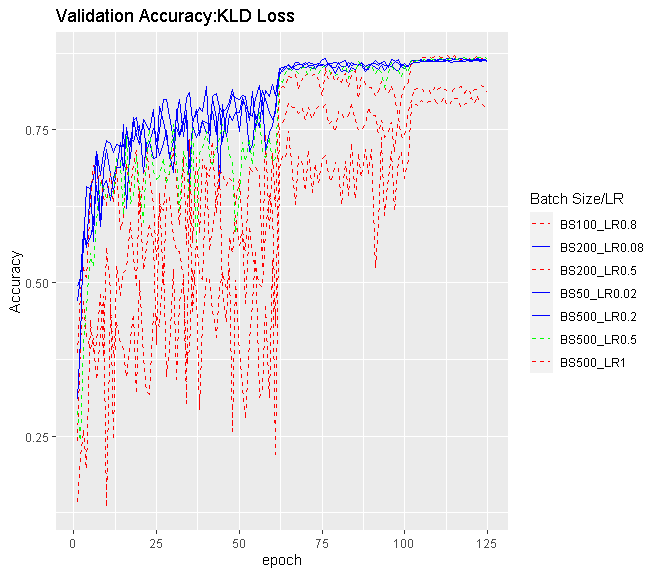}
\end{figure}

\subsection{Equation \cite{W2019}(4.45) for KLD Loss}

In Figure \ref{KLDLossGraphs}, we use the Kullback-Leibler Divergence (KLD) as our loss function, scale both the learning rate and batch size, and compare them against rescaling the batch size exactly with the learning rate. The blue lines represent experiments where the BS/LR is kept the same, i.e., rescaling the batch size and learning rate by the same amount. We notice that the evolution of their curves in both the training and validation accuracy very closely match. The green line represents a BS/LR ratio that is relatively close to those of the blue lines, and we can see that the evolution of its curve is somewhat close to that of the blue lines. Finally, the red lines represent BS/LR ratios that are relatively far from those of the blue and green lines; we can see that evolution of the learning curves of those red lines is noticeably different from that of the blue and green lines. These results numerically confirm the validity of Equation \cite{W2019}(4.45).

\begin{figure}[h!]
\caption{Evolution of Training and Validation Accuracy for varying BS/LR using ResNet-56 on CIFAR-10 with Gradient Log Loss.}
\label{GradientLogLossGraphs}
\centering
\includegraphics[scale=0.63]{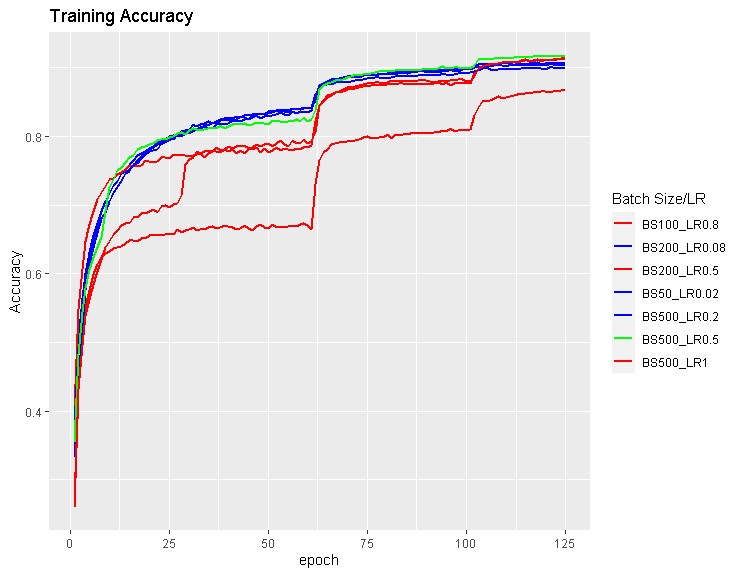}
\includegraphics[scale=0.6]{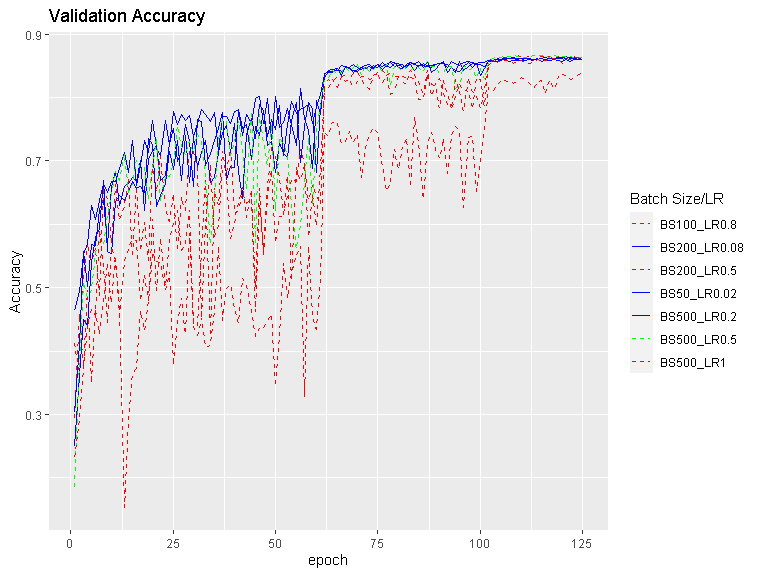}
\end{figure}

\subsection{Equation \cite{W2019}(4.46) for Gradient Log Loss}

In Figure \ref{GradientLogLossGraphs}, we use the gradient of the log loss as our loss function, scale both the learning rate and batch size, and compare them against rescaling the batch size exactly with the learning rate. The blue lines represent experiments where the BS/LR is kept the same, i.e. rescaling the batch size and learning rate by the same amount. We notice that the evolution of their curves in both the training and validation accuracy very closely match. The green line represents a BS/LR ratio that is relatively close to those of the blue lines, and we can see that the evolution of its curve is somewhat close to that of the blue lines. Finally, the red lines represent BS/LR ratios that are relatively far from those of the blue and green lines; we can see that evolution of the learning curves of those red lines is noticeably different from that of the blue and green lines. These results numerically confirm the validity of Equation \cite{W2019}(4.46).

\subsection{ Equation \cite{W2019}(4.45) vs. Equation \cite{J2018}(9) for KLD Loss}

We next analyze the values of $tr[{\bm \sigma}^2(X(\infty))]$ and $\text{Tr}({\bf H})$ from Equations \cite{W2019}(4.45) and \cite{J2018}(9), respectively, for varying batch sizes and learning rates under the KLD loss. We take 5 runs for each of the experiments and average the resulting values of $tr[{\bm \sigma}^2(X(\infty))]$ and $\text{Tr}({\bf H})$. The results are presented in Table \ref{445vs9}.

\begin{table}
\centering
\begin{tabular}{lr*{3}{c}rr}
Experiment & BS/LR      & BS & LR &  tr$({\bf H})$ & $tr[{\bm \sigma}^2(X(\infty))]$  & Magnitude Difference \\
\hline
 1&1,250 &100 & 0.08 & 22,398,330 & 9,528,207 & 2.4\\
  2& 500   &100& 0.2 & 44,428,697& 5,501,901 &8.1  \\
    4& 200   &100& 0.5 & 45,588,319  &  4,575,870& 9.9 \\ 
 4&2,500 &200 & 0.08 & 31,897,024 & 10,304,980 & 3.1 \\
 5& 1,000 &200&0.2 &33,932,327 & 4,084,640 & 8.3  \\
 6&400 &200 & 0.5 & 41,058,846 & 3,486,877 & 11.8 \\
  7&6,250 &500 & 0.08 & 63,343,355 & 9,530,501 & 6.6\\
  8&2,500  &500 & 0.2 & 35,252,033  & 4,261,116  & 8.3  \\
    9&1,000  &500 & 0.5 & 54,212,278   & 5,380,719  &  10.1 
\end{tabular}
\caption{Values of $tr[{\bm \sigma}^2(X(\infty))]$ and $\text{Tr}({\bf H})$ for varying batch sizes and learning rates under KLD loss, averaged over 5 runs for each experiment. Magnitude Difference represents $\frac{\text{Tr}({\bf H})}{tr[{\bm \sigma}^2(X(\infty))]}.$}
 \label{445vs9}
\end{table}

\begin{figure}[h!]
\caption{Evolution of Magnitude Difference in $\text{Tr}({\bf H})$ and $tr[{\bm \sigma}^2(X(\infty))]$ for varying Batch Sizes and Learning Rates.}
\label{MagDiff}
\centering
\includegraphics[scale=0.6]{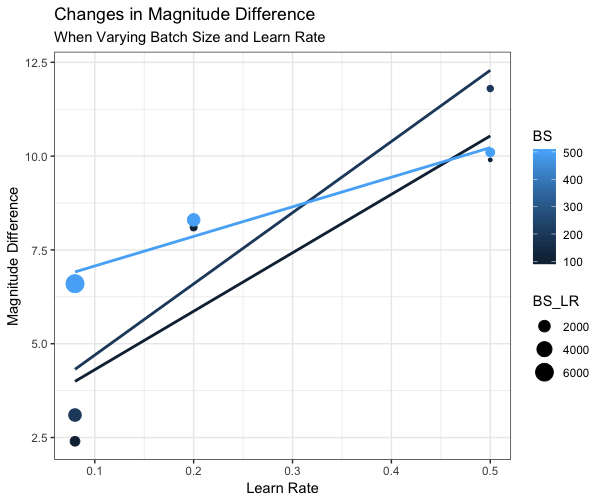}
\includegraphics[scale=0.6]{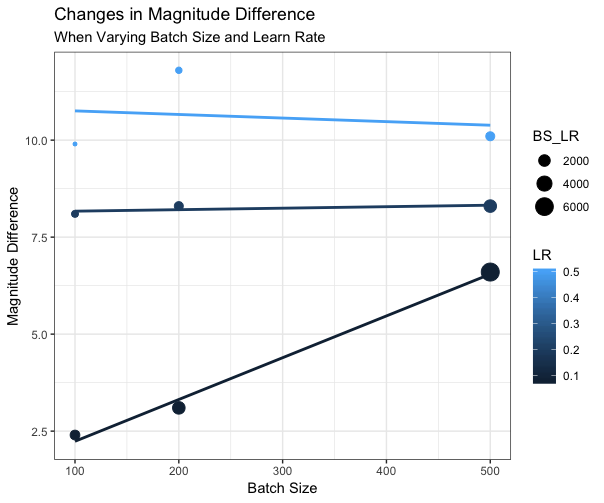}
\end{figure}

Generally, it seems that if batch size is held constant, a higher learning rate causes $\text{Tr}({\bf H})$ to increase, while $tr[{\bm \sigma}^2(X(\infty))]$ decreases. If learning rate is held constant, as batch size increases, $tr[{\bm \sigma}^2(X(\infty))]$ seems to hold constant, while $\text{Tr}({\bf H})$ increases. In addition, as the BS/LR ratio decreases, it appears that the magnitude difference in $\text{Tr}({\bf H})$ and $tr[{\bm \sigma}^2(X(\infty))]$ generally increases. It appears that changes in learning rate affect the magnitude difference much stronger than changes in batch size; holding batch size constant while increasing the learning rate leads to larger magnitude differences, while holding learning rate constant and increasing the batch size doesn't affect the magnitude difference as much. These observations are illustrated in Figure \ref{MagDiff}. In all cases, we notice that $tr[{\bm \sigma}^2(X(\infty))]$ is magnitudes smaller than $\text{Tr}({\bf H})$, suggesting that Equation \cite{W2019}(4.45) is a more robust formulation of the the relationship between expected loss and BS/LR ratio compared to Equation \cite{J2018}(9).

We note that similar experiments were conducted under log loss, and Equations \cite{W2019}(4.45) and \cite{J2018}(9) seem to generally agree in that case, due to the assumptions made in Jastrz\c{e}bski et al. (2018). 

\section*{Acknowledgments}
We would like to acknowledge support for this project from the National Science Foundation
(NSF grants DMS-1707605 and DMS-1913149).

\newpage



\begin{thebibliography}{9}

\bibitem{BrownBarth1989}
A. A. Brown and M. C. Bartholomew-Biggs, 
\textit{Some Effective Methods for Unconstrained Optimization Based on the Solution of Systems of Ordinary Differential Equations}, 
Journal of Optimization Theory and Applications, 62(2):211-224, 1989.


\bibitem{}
D. J. Foster, A. Sekhari, O. Shamir, N. Srebro, K. Sridharan, B. Woodworth, 
\textit{The Complexity of Making the Gradient Small in Stochastic Convex Optimization}, 
Proceedings of Machine Learning Research 99, 1-27 (2019).

\bibitem{G2009}
C. W. Gardiner,
\textit{Stochastic Methods: A Handbook for the Natural and Social Sciences},
Springer, 4th edition. 2009.


\bibitem{J2018}
S. Jastrz\c{e}bski, Z. Kenton, D. Arpit, N. Ballas, A. Fischer, Y. Bengio, and A. Storkey. 
\textit{Three Factors Influencing Minima in SGD},
September 14, 2018. arXiv: 1711.04623v3 [cs.LG].

\bibitem{Kawaguchi}
K. Kawaguchi,
\textit{Deep Learning Without Poor Local Minima}, 
In Advances In Neural Information Processing Systems, pages 586-594, 2016.


\bibitem{Nesterov}
Y. Nesterov, 
\textit{Gradient Methods for Minimizing Composite Functions}, 
Mathematical Programming, 140(1):125-161, 2013.

\bibitem{Ruder}
S. Ruder,
\textit{An Overview of Gradient Descent Optimization Algorithms}, arXiv:1609.04747v1, 2016 [cs.LG].

\bibitem{W2019}
Y. Wang,
\textit{Asymptotic Analysis via Stochastic Differential Equations of Gradient Descent Algorithms in Statistical and Computational Paradigms},
November 12, 2019. arXiv:1711.09514v5 [stat.ML].  A version co-authored with Shang Wu 
will be published in Journal of Machine Learning Research 21, 2020.  
\end{thebibliography}
\end{document}